\documentclass[11pt]{tibop-article}
\TIBauthor[Hanna Abi Akl]{Hanna Abi Akl\affOne\affTwo\orcidlink{0000-0001-9829-7401}}
\TIBaffiliations{\affOne Data ScienceTech Institute (DSTI), 4 Rue de la Collégiale, 75005, Paris, France\medskip\\
	\affTwo Université Côte d’Azur, Inria, CNRS, I3S\medskip\\
	*Correspondence: Hanna Abi Akl, hanna.abi-akl@dsti.institute
	
}

\TIBtitle{DSTI at LLMs4OL 2024 Task A: Intrinsic versus extrinsic knowledge for type classification}
\TIBsubtitle{Applications on WordNet and GeoNames datasets}
\TIBbundlename[Open Conf Proc 4 (2024) "LLMs4OL 2024: The 1st Large Language Models for Ontology Learning Challenge at the 23rd ISWC"]{LLMs4OL 2024: The 1st Large Language Models for Ontology Learning Challenge at the 23rd ISWC}
\TIBabstract{We introduce semantic towers, an extrinsic knowledge representation method, and compare it to intrinsic knowledge in large language models for ontology learning. Our experiments show a trade-off between performance and semantic grounding for extrinsic knowledge compared to a fine-tuned model's intrinsic knowledge. We report our findings on the Large Language Models for Ontology Learning (LLMs4OL) 2024 challenge.}
\TIBkeywords{Large Language Models, Ontology Learning, Semantic Web, Knowledge Representation, Semantic Primes}
\usepackage{xfrac}
\begin{document}
\maketitle

\section{Introduction and related work}
\label{section-introduction}
Large language models (LLMs) have seen widespread applications across different tasks in the fields of Natural Language Processing and Knowledge Representation. Particularly, LLM-based systems are used to tackle ontology-related tasks such as ontology learning \cite{ronzano2024towards}, knowledge graph construction \cite{kommineni2024human}, ontology matching \cite{giglou2024llms4om}\cite{he2023exploring} and ontology generation \cite{toro2023dynamic}. Retrieval-Augmented-Generation (RAG) systems, which build on the capabilities of LLMs by enhancing retrieval using external knowledge sources, have also shown promising results in tasks involving the use of ontologies \cite{buehler2024generative}. On the other hand, symbolic methods like semantic representation using primes and universals \cite{wierzbicka1996semantics} form another research frontier in the area of knowledge representation which is at the heart of ontologies \cite{fahndrich2018semantic}.
\par
In this work, we evaluate and compare the performance of fine-tuned models on Task A of the LLMs4OL \cite{babaei2023llms4ol}\cite{llms4ol2024overview}\cite{llms4ol2024dataset} 2024 challenge\footnote{https://sites.google.com/view/llms4ol/home} using intrinsic LLM knowledge and external knowledge sources we define as semantic towers. The rest of the work is organized as follows. In section
\ref{section-methodology}, we present our methodology. Section \ref{section-experiments} describes our experimental framework. In section \ref{section-results}, we report our results and discuss our findings. Finally, we conclude in section \ref{section-conclusion}.

\section{Methodology}
\label{section-methodology}
This section describes the methodology for creating a semantic tower \textit{ST} which we define as:
\begin{equation}
    ST = \{s_1,s_2,..,s_n\},
\end{equation}
where \textit{s} is a domain semantic primitive pointing to a semantic property for a given domain and \textit{n} is the minimal number of primitives needed to define the domain. The rest of this section details the construction of domain semantic towers from semantic primitives.

\subsection{Domain semantic primitives}
For each domain, we use the Wikidata Query Service\footnote{https://query.wikidata.org/} to retrieve semantic information for each term type category. This body of information, or semantic set, serves as the base for the domain semantic primitives.    \par
The WordNet semantic set consists of: \{subclass,instance,part,represents,description\}. The GeoNames semantic set consists of: \{subclass,instance,part,category,description\}.

\subsection{Semantic towers}
The construction scheme of semantic towers is domain-invariant and summarized in the following steps:  \par

\begin{enumerate}
    \item The values of the semantic set for each term type are tokenized into a bag of words, cleaned and normalized through lowercase transformation and stop word removal.
    \item The result is transformed to a comma-separated list.
    \item Empty values and duplicates are pruned from the list.
    \item The list of primitives is transformed to vector embeddings of size 1024 using the gte-large\footnote{https://huggingface.co/thenlper/gte-large} model by Google \cite{li2023towards}.
    \item The resulting domain vector embeddings are stored in a MongoDB\footnote{https://www.mongodb.com/} collection to form a vector store, i.e. the semantic tower.
    \item The semantic tower is indexed on embeddings search for optimized performance.
\end{enumerate}

\begin{figure}[h!]
    \includegraphics[width=.99\linewidth]{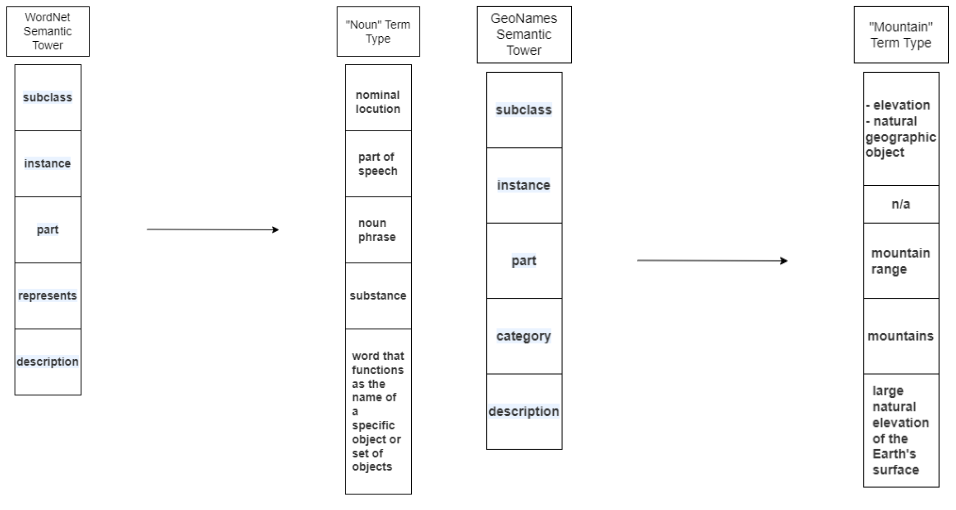}
    \caption{WordNet and GeoNames semantic towers with examples.}\label{wordnet-geonames-semantic-tower}
\end{figure}

Figure \ref{wordnet-geonames-semantic-tower} shows examples of the WordNet and GeoNames semantic towers.

\section{Experiments}
\label{section-experiments}
This section describes our experiments in terms of data, models and training process.

\subsection{Dataset description} 
We consider two datasets for our experiments: WordNet and GeoNames. Both datasets are used for training and testing our models in the respective subtasks (A.1 and A.2). The dataset descriptions are detailed in the following subsections.

\subsubsection{WordNet} 
\begin{figure}[h!]
    \includegraphics[width=.6\linewidth]{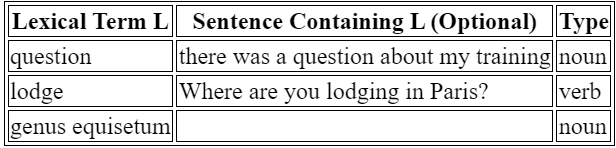}
    \caption{Subtask A.1 term typing WordNet examples.}\label{wordnet-data-example}
\end{figure}

The dataset consists of 40,559 train terms and 9,470 test terms. It contains four types to classify each term: noun, verb, adjective, adverb. Figure \ref{wordnet-data-example} shows example data.

\subsubsection{GeoNames}
\begin{figure}[h!]
    \includegraphics[width=.3\linewidth]{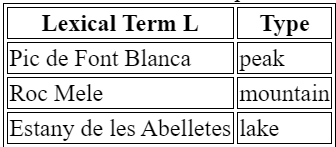}
    \caption{Subtask A.2 term typing GeoNames examples.}\label{geonames-data-example}
\end{figure}

The dataset consists of 8,078,865 train terms and 702,510 test terms. It contains 660 categories of geographical locations. Example data is presented in Figure \ref{geonames-data-example}.

\subsection{System description}
This section describes the models as well as the setup of our experiments.

\subsubsection{Models}
We train one model for each subtask. We use the same base flan-t5-small\footnote{https://huggingface.co/google/flan-t5-small} model and fine-tune it on the subtask datasets respectively. The training hyperparameters for both models are configured identically: \{learning\_rate: 1e-05, train\_batch\_size: 4, eval\_batch\_size: 4, num\_epochs: 5, question\_length: 512, target\_length: 512, optimizer: Adam\}. For subtask A.1, the model is trained on 70\% of the provided WordNet dataset and the remaining 30\% is used for validation. Table \ref{tab:flan-wordnet-train-results} shows the training results. \par

\begin{table}[h!]
    \caption{Subtask A.1 model training results.}\label{tab:flan-wordnet-train-results}
    \begin{tabularx}{\linewidth}{|X|X|X|X|}
        \hline
        \textbf{Training Loss}    & \textbf{Epoch} & \textbf{Step} & \textbf{Validation Loss}                            \\
        \hline
        0.1725              & 1.0 & 1000 & 0.0640                  \\
        \hline
        0.1250              & 2.0 & 2000 & 0.0535                  \\
        \hline
        0.1040              & 3.0 & 3000 & 0.0469 \\
        \hline
        0.0917              & 4.0 & 4000 & 0.0421        \\
        \hline
        0.0830              & 5.0 & 5000 & 0.0384        \\
        \hline
    \end{tabularx}
\end{table}

For subtask A.2, the length of the data makes fine-tuning challenging. To remedy this problem, we curate a subset from the original dataset using the following algorithm:

\begin{enumerate}
    \item Each type category is counted into a length variable \textit{cat\_len}.
    \item For each category represented less than 100 times (i.e. \textit{cat\_len $<$ 100}), all terms classified in that category are selected and kept in the dataset.
    \item If \textit{cat\_len $\geq$ 100}, only the first 25 terms classified in that category are selected. The threshold of 25 keeps the size of the dataset relatively small given the large number of categories.
\end{enumerate}

We obtain a curated dataset of 2041 terms representing all possible categories. The model is trained on 70\% of the curated dataset and the remaining 30\% is used for validation. Table \ref{tab:flan-geonames-train-results} shows the training results.

\begin{table}[h!]
    \caption{Subtask A.2 model training results.}\label{tab:flan-geonames-train-results}
    \begin{tabularx}{\linewidth}{|X|X|X|X|}
        \hline
        \textbf{Training Loss}    & \textbf{Epoch} & \textbf{Step} & \textbf{Validation Loss}                            \\
        \hline
        2.6223 &	1.0 &	1000 &	1.5223                  \\
        \hline
        2.1430 &	2.0 &	2000 &	1.3764                  \\
        \hline
        1.9100 &	3.0 &	3000 &	1.2825 \\
        \hline
        1.7642 &	4.0 &	4000 &	1.2102        \\
        \hline
        1.6607 &	5.0 &	5000 &	1.1488        \\
        \hline
    \end{tabularx}
\end{table}

The training of both models is done on a Google Colab instance using an A100 High-RAM GPU. Both A.1 and A.2 models are available publicly on Hugging Face respectively under the names flan-t5-small-wordnet\footnote{https://huggingface.co/HannaAbiAkl/flan-t5-small-wordnet} and flan-t5-small-geonames\footnote{https://huggingface.co/HannaAbiAkl/flan-t5-small-geonames}.

\subsubsection{Features}
The same feature engineering method is applied for both models. It consists in embedding input text into vectors of size 1024 using the gte-large model. For the flan-t5-small-wordnet model, the input is the concatenation of the term and the sentence when provided. For flan-t5-small-geonames, the input text is the term.

\subsubsection{Setup}
We conduct two experiments per subtask for a total of four.
\par
For subtask A.1, the first experiment (WN1) consists in prompting the fine-tuned WordNet model on the test split of the provided dataset which is used as an unofficial test set ahead of the official submission. The prompt used for the model is: \textbf{\textit{Give the entity for the term X. Select the answer from this list Y}}, where X is dynamically replaced by the input term and Y is replaced by the list of possible term types.
    
\begin{figure}[h!]
    \includegraphics[width=.6\linewidth]{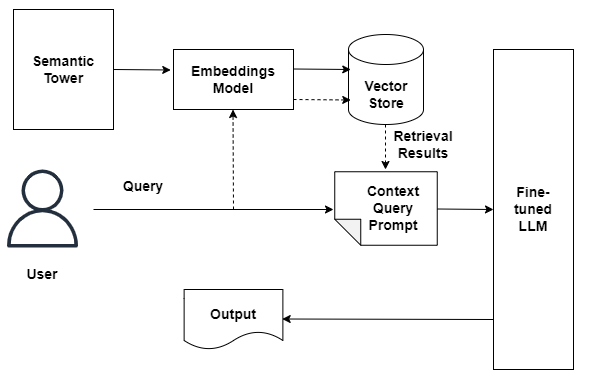}
    \caption{RAG system architecture.}\label{rag-system-architecture}
\end{figure}

The second experiment (WN2) leverages the RAG pipeline shown in Figure \ref{rag-system-architecture} in conjunction with a user prompt to retrieve the best term type for each input term. The input is vectorized and compared to the embeddings of the WordNet semantic tower for each term type. A cosine similarity score is used to determine the closest type from the semantic tower vector store to return the top 1 candidate. The answer is then used as an additional input to the user prompt given to the model: \textbf{\textit{Give the entity for the term X. Select the answer from this list Y relying on the search result Z}}, where X and Y are as previously defined and Z represents the best-matched term type from the semantic tower.

For subtask A.2, both experiments GN1 and GN2 mimic WN1 and WN2 respectively. For GN1, the fine-tuned GeoNames model is evaluated on the test split of the curated dataset. The user prompt for the model is the same as that of WN1, with the only changes being the X term values and the Y list of types which now refers to the geographical categories.

In experiment GN2, the same pipeline from Figure \ref{rag-system-architecture} is reproduced with the only difference being the replacement of the WordNet semantic tower with the GeoNames semantic tower. The user prompt used for the fine-tuned model is the same as that of WN2, with the Y list reflecting the geographical categories. All experiments are conducted on a Google Colab instance using a L4 High-RAM GPU. The code for our experimental setup is publicly available on GitHub\footnote{https://github.com/HannaAbiAkl/SemanticTowers}.

\section{Results}
\label{section-results}
Table \ref{tab:wn-test-results} shows our experimental results on the WordNet test set. The results of the GeoNames experiments are presented in Table \ref{tab:gn-test-results}. The F1 scoring metric reflects the criteria of performance assessment set by the task organizers.\par
Experiments WN1 and GN1 perform better than WN2 and GN2 respectively, with a performance gain close to 10\%. At first inspection, the results seem to suggest that the flan-t5 model, with a little fine-tuning, can rely on its existing knowledge regarding the dataset domains to correctly classify terms by type. The use of an external knowledge base, such as a semantic tower, seems to create more errors in the model answers. However, closer examination of a subset of the outputs reveals that semantic towers effectively ground certain semantic notions in the model that are otherwise lost if the model only relies on its existing knowledge. Examples include correctly classifying the term \textit{into the bargain} as \textit{adverb} with the aid of the WordNet semantic tower (as opposed to classifying it as \textit{noun} without it). While the word \textit{bargain} dominates the term in the example, the flan-t5-small-wordnet model misses out on the correct classification which attributes an important weight to the adverb \textit{into} that becomes more prominent with the semantic tower embeddings representation. A similar case can be made for the GeoNames experiments, where the usage of the semantic tower in conjunction with the model improves the classification choice for plural categories (e.g. terms classified as \textit{mountains}, \textit{peaks}, \textit{streams}). The outputs of experiment GN1 show that the model alone has a tendency to choose the singular forms of these categories which count for incorrect classifications. Moreover, experiment GN2 also shows that the semantic tower helps ground nuances between categories (e.g. \textit{stream} versus \textit{section of stream}) which leads to a more fine-grained (and accurate) typing.

For the official test sets released by the task organizers, we evaluate only the A.1 subtask using WN1 and WN2 and present our results in Table \ref{tab:wn-submission}. Both WN1 and WN2 demonstrate a slight drop in performance of around 1\% but perform competitively well. The results demonstrate that the model training as well as the WordNet semantic tower construction are sound enough to avoid catastrophic drift.

We refrain from submitting to the other subtasks, most notably A.2, because of the length of the official test set which is extremely challenging to run on our available resources.

\begin{table}[h!]
    \caption{Experimental results on the WordNet set.}\label{tab:wn-test-results}
    \begin{tabularx}{\linewidth}{|X|X|}
        \hline
        \textbf{Experiment}    & \textbf{F1}                             \\
        \hline
        \textbf{flan-t5-small-wordnet (WN1)}              & \textbf{0.9820}                  \\
        \hline
        flan-t5-small-wordnet + WordNet semantic tower (WN2)              & 0.8581                   \\
        \hline
    \end{tabularx}
\end{table}

\begin{table}[h!]
    \caption{Experimental results on the GeoNames set.}\label{tab:gn-test-results}
    \begin{tabularx}{\linewidth}{|X|X|}
        \hline
        \textbf{Experiment}    & \textbf{F1}                             \\
        \hline
        \textbf{flan-t5-small-geonames (GN1)}              & \textbf{0.6820}                  \\
        \hline
        flan-t5-small-geonames + GeoNames semantic tower (GN2)              & 0.5636                   \\
        \hline
    \end{tabularx}
\end{table}

\begin{table}[h!]
    \caption{Subtask A.1 (few-shot) WordNet term typing leaderboard.}\label{tab:wn-submission}
    \begin{tabularx}{\linewidth}{|X|X|X|X|}
        \hline
        \textbf{Teal Name}    & \textbf{F1} & \textbf{Precision} & \textbf{Recall}                            \\
        \hline
        TSOTSALearning              & 0.9938  & 0.9938  & 0.9938  \\
        \hline
        \textbf{DSTI (WN1)}              & \textbf{0.9716} & \textbf{0.9716} & \textbf{0.9716}                  \\
        \hline
        DaseLab              & 0.9697 & 0.9689 & 0.9704                  \\
        \hline
        RWTH-DBIS              & 0.9446  & 0.9446  & 0.9446  \\
        \hline
        TheGhost              & 0.9392 & 0.9389 & 0.9395        \\
        \hline
        Silp\_nlp              & 0.9037 & 0.9037 & 0.9037        \\
        \hline
        \textbf{DSTI (WN2)}              & \textbf{0.8420} & \textbf{0.8420} & \textbf{0.8420}         \\
        \hline
        Phoenixes              & 0.8158 & 0.7689 & 0.8687 \\
        \hline
    \end{tabularx}
\end{table}

\section{Conclusion}
\label{section-conclusion}
In this shared task, we investigate and compare intrinsic knowledge in LLMs with external semantic sources for ontology learning. While the introduction of semantic towers proves there is still some way to go to achieve semantic resonance in LLMs, it shows promising results in grounding these models semantically and fine-graining their knowledge. Our fine-tuned models demonstrate that ontology term typing is a task within the reach of LLMs based on their existing knowledge. In future work, we will explore the potential of semantic towers and expand their implementation to existing LLM-based systems. \par

\printbibliography[heading=references]

@article{llms4ol2024overview,
         title={LLMs4OL 2024 Overview: The 1st Large Language Models for Ontology Learning Challenge},
         volume={4},
         journal={Open Conference Proceedings},
         author={Babaei Giglou, Hamed
                 and D'Souza, Jennifer
                 and Auer, S{\"o}ren},
         year={2024},
         month={Oct.}
}

@article{llms4ol2024dataset,
         title={LLMs4OL 2024 Datasets: Toward Ontology Learning with Large Language Models},
         volume={4},
         journal={Open Conference Proceedings},
         author={Babaei Giglou, Hamed
                 and D'Souza, Jennifer
                 and Sadruddin, Sameer
                 and Auer, S{\"o}ren},
         year={2024},
         month={Oct.}
}

@inproceedings{babaei2023llms4ol,
  title={LLMs4OL: Large language models for ontology learning},
  author={Babaei Giglou, Hamed and D’Souza, Jennifer and Auer, S{\"o}ren},
  booktitle={International Semantic Web Conference},
  pages={408--427},
  year={2023},
  organization={Springer}
}

@article{buehler2024generative,
  title={Generative retrieval-augmented ontologic graph and multiagent strategies for interpretive large language model-based materials design},
  author={Buehler, Markus J},
  journal={ACS Engineering Au},
  volume={4},
  number={2},
  pages={241--277},
  year={2024},
  publisher={ACS Publications}
}

@book{fahndrich2018semantic,
  title={Semantic decomposition and marker passing in an artificial representation of meaning},
  author={F{\"a}hndrich, Johannes},
  year={2018},
  publisher={Technische Universitaet Berlin (Germany)}
}

@article{giglou2024llms4om,
  title={LLMs4OM: Matching Ontologies with Large Language Models},
  author={Giglou, Hamed Babaei and D'Souza, Jennifer and Auer, S{\"o}ren},
  journal={arXiv preprint arXiv:2404.10317},
  year={2024}
}

@article{he2023exploring,
  title={Exploring large language models for ontology alignment},
  author={He, Yuan and Chen, Jiaoyan and Dong, Hang and Horrocks, Ian},
  journal={arXiv preprint arXiv:2309.07172},
  year={2023}
}

@article{kommineni2024human,
  title={From human experts to machines: An LLM supported approach to ontology and knowledge graph construction},
  author={Kommineni, Vamsi Krishna and K{\"o}nig-Ries, Birgitta and Samuel, Sheeba},
  journal={arXiv preprint arXiv:2403.08345},
  year={2024}
}

@article{li2023towards,
  title={Towards general text embeddings with multi-stage contrastive learning},
  author={Li, Zehan and Zhang, Xin and Zhang, Yanzhao and Long, Dingkun and Xie, Pengjun and Zhang, Meishan},
  journal={arXiv preprint arXiv:2308.03281},
  year={2023}
}

@article{ronzano2024towards,
  title={Towards Ontology-Enhanced Representation Learning for Large Language Models},
  author={Ronzano, Francesco and Nanavati, Jay},
  journal={arXiv preprint arXiv:2405.20527},
  year={2024}
}

@article{toro2023dynamic,
  title={Dynamic retrieval augmented generation of ontologies using artificial intelligence (DRAGON-AI)},
  author={Toro, Sabrina and Anagnostopoulos, Anna V and Bello, Sue and Blumberg, Kai and Cameron, Rhiannon and Carmody, Leigh and Diehl, Alexander D and Dooley, Damion and Duncan, William and Fey, Petra and others},
  journal={arXiv preprint arXiv:2312.10904},
  year={2023}
}

@book{wierzbicka1996semantics,
  title={Semantics: Primes and universals: Primes and universals},
  author={Wierzbicka, Anna},
  year={1996},
  publisher={Oxford University Press, UK}
}

\end{document}